\documentclass[11pt,a4paper]{article}
\usepackage[hyperref]{acl2021}
\usepackage{times}
\usepackage{latexsym}

\usepackage{microtype}

\usepackage{graphicx}
\usepackage{amsmath}
\usepackage{amsfonts}
\usepackage[noend]{algpseudocode}
\usepackage{algorithmicx,algorithm}
\usepackage{booktabs}
\usepackage{multirow}
\usepackage{color,xcolor}
\usepackage{soul}

\aclfinalcopy 
\def\aclpaperid{609} 

\title{BoB: BERT Over BERT for Training Persona-based Dialogue Models from Limited Personalized Data}

\author{Haoyu Song$^1$, Yan Wang, Kaiyan Zhang$^1$, Wei-Nan Zhang$^{1}$\thanks{\ \ Wei-Nan Zhang is the corresponding author.}, Ting Liu$^1$ \\
  $^1$Research Center for Social Computing and Information Retrieval\\
  Harbin Institute of Technology, Heilongjiang, China\\
  \texttt{\{hysong,kyzhang,wnzhang,tliu\}@ir.hit.edu.cn}\\
  \texttt{yanwang.branden@gmail.com} \\
  }

\date{}

\begin{document}
\maketitle

\begin{abstract}

Maintaining consistent personas is essential for dialogue agents. Although tremendous advancements have been brought, the limited-scale of annotated persona-dense data are still barriers towards training robust and consistent persona-based dialogue models. In this work, we show how the challenges can be addressed by disentangling persona-based dialogue generation into two sub-tasks with a novel BERT-over-BERT (BoB) model. Specifically, the model consists of a BERT-based encoder and two BERT-based decoders, where one decoder is for response generation, and another is for consistency understanding. In particular, to learn the ability of consistency understanding from large-scale non-dialogue inference data, we train the second decoder in an unlikelihood manner. Under different limited data settings, both automatic and human evaluations demonstrate that the proposed model outperforms strong baselines in response quality and persona consistency.

\end{abstract}

\section{Introduction}

Various approaches have been explored to introduce explicit personas in dialogue models~\cite{ijcai2018-595,ijcai2019-721,Zheng_Zhang_Huang_Mao_2020,liu-etal-2020-impress}. The PERSONA can be defined as a composite of elements of identity, such as profiles and background personal facts.
In persona-based dialogues, the generated responses are conditioned not only on the dialogue context but also on some predefined personas, so the presenting personality could be more consistent.

\begin{figure}[t]
\centering
\includegraphics[width=.99\columnwidth]{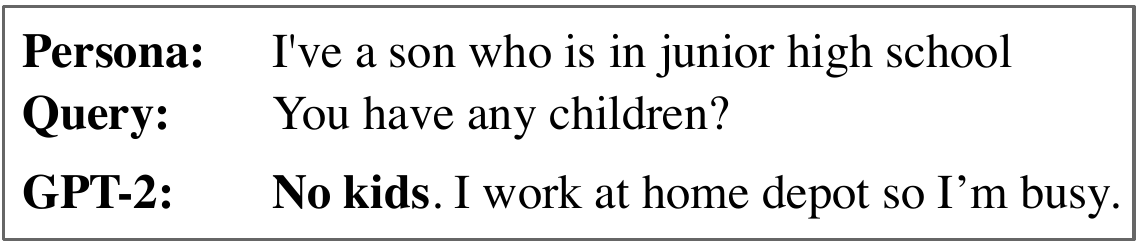}
\caption{ A 12-layer GPT2 finetuned on PersonaChat dataset still generates an inconsistent response. }
\label{fig:1}
\end{figure}

Existing persona-based dialogue models heavily utilize a set of persona-related dialogue data~\cite{wolf2019transfertransfo,golovanov2020lost}, such as the PersonaChat~\cite{zhang-etal-2018-personalizing}. This kind of crowd-sourced dataset covers rich persona features, namely ``persona-dense''.
Nevertheless, the scale of such crowd-sourced datasets is limited by the expensive costs: two annotators are asked to act the part of a given provided persona and chat naturally to get to know each other during the conversation.
On the other hand, conversations in daily life are not always persona-related. According to Twitter content analysis, less than 10\% messages on Twitter reveal personal anecdote or activities at home or work and even less for personally identifiable information~\cite{naaman2010really,humphreys2014twitter}. As a result, the large-scale data collected from social media would only contain a limited amount of persona-related dialogues, which is ``persona-sparse''. The limited-scale of crowd-sourced data and the persona-sparsity in large-scale data present one common challenge: a model trained on limited personalized data cannot sufficiently understand persona consistency. As shown in Figure~\ref{fig:1}, a 12-layer GPT2~\cite{radford2019language} finetuned on the PersonaChat dataset still shows a lack of consistency.

After rethinking the essence of persona-based dialogue generation, we can find that it requires the dialogue agent to own the capabilities to 1) understand the persona-response consistency and 2) generate a persona-related response given the dialogue context. Obviously, an ideal dataset that satisfies both features are difficult to annotate. However, once we disentangle persona-based dialogue generation into two sub-tasks: consistency understanding and dialogue generation, it is easy to find abundant data resources for them. For consistency understanding, we may leverage large-scale non-dialogue inference data, such as SNLI~\cite{bowman-etal-2015-snli} and MNLI~\cite{williams-etal-2018-mnli} as the training data. As for dialogue generation, we already have various large-scale persona-sparse datasets.  
 
Inspired by the aforementioned motivation, in this work, we explore to learn a consistent persona-based dialogue model from limited personalized dialogues, with the assistance of large-scale non-dialogue inference data. Specifically, the proposed model consists of an encoder $\mathbb{E}$, an auto-regressive decoder $\mathbb{D}_1$ for response generation, and a bidirectional decoder $\mathbb{D}_2$ for consistency understanding. Given personas $P$ and dialogue query $Q$, the $\mathbb{E}$ and $\mathbb{D}_1$ jointly work in an encoder-decoder manner to capture a typical query to response mapping $F_G(S|Q,P)$, and generate a coarse response representation $R_1$. Then $R_1$ and personas $P$ are fed into the bidirectional decoder $\mathbb{D}_2$ to map $R_1$ to final response representations $R_2$: $F_U(R_2|S,P)$. Since the consistency understanding part $F_U(R|S,P)$ is independent of the dialogue query $Q$, it can be learned on non-dialogue inference datasets. Here an unlikelihood training objective~\cite{welleck2019neural} is applied to make contradicted cases in the inference data less likely so that $\mathbb{D}_2$ could acquire the ability of consistency understanding.

We initialize all modules from BERT~\cite{devlin-etal-2019-bert} and name the proposed model BERT-over-BERT (BoB). To verify the effectiveness of our model, we experiment on two limited data scenarios: 1) a persona-dense scenario~\cite{zhang-etal-2018-personalizing} with low-resource settings~\cite{zhao2019low}, and 2) a persona-sparse scenario~\cite{Zheng2019PersonalizedDG}. Both automatic and human evaluations indicate that our model generalizes well under different settings and outperforms strong baselines on most metrics, especially on persona consistency.

Contributions in this work are three-fold:
\begin{itemize}
  \item We disentangled the task of persona-based dialogue generation into two sub-tasks: consistency understanding and dialogue generation.
  \item A BERT-based generative framework, BoB, was proposed for training persona-based dialogue models from limited data.
  \item An unlikelihood training method with non-dialogue inference data was introduced to enhance persona consistency understanding.
\end{itemize}

\section{Related Work}

\paragraph{Persona-based Dialogues}
Recent studies on persona-based dialogue generation focus on a data-driven manner. They learn persona-related features directly from personalized dialogue datasets, either with implicit persona embeddings~\cite{li-etal-2016-persona} or with explicit profiles~\cite{ijcai2018-595} and personal facts~\cite{mazare-etal-2018-training}. Following this research line, more sophisticated neural models are emerging, such as modeling mutual-persona~\cite{liu-etal-2020-impress} and multi-stage persona-based dialogue generation~\cite{song-etal-2020-generate}.

Meanwhile, various pre-training methods have also been applied in this field. \citet{wolf2019transfertransfo} and~\citet{golovanov2020lost} show that fine-tuning pre-trained GPT on the persona-dense dataset can improve the quality of generated responses. \citet{Zheng_Zhang_Huang_Mao_2020} propose an attention-routing mechanism in a GPT-based model to control the flow of persona information. \citet{lin2020xpersona} explore how to leverage BERT model for dialogue generation. Different large-scale pretrained chatbots~\cite{roller2020recipes,madotto2020adapter} also show their effectiveness on persona-based dialogues.

\paragraph{Disentangled Representation} The concept of ``disentangling'' can be defined as transformations that only change some properties of the underlying model while leaving all other properties invariant~\cite{DBLP:journals/corr/abs-1812-02230}. 
The variational autoencoder~\cite{kingma2013auto} could be regarded as a disentangled representation learning framework, and various methods are built within it~\cite{kim2018disentangling,locatello2019disentangling}. 

\paragraph{Unlikelihood Training}
Likelihood tries to maximize the probability of target sequence, while unlikelihood corrects known biases by minimizing the probability of negative candidates~\cite{welleck2019neural}. Closely related to our work, \citet{li-etal-2020-dont} first explored unlikelihood training in addressing dialogue logical contradictions. They get contradicted dialogues from PersonaChat according to DNLI~\cite{welleck-etal-2019-dialogue}, a PersonaChat-oriented dialogue inference dataset. Then unlikelihood training is applied to reduce the probability of contradicted responses. Different from~\citet{li-etal-2020-dont}, with carefully designed decoders, our model could learn from large-scale non-dialogue inference datasets, making it generalizable to different scenarios, such as persona-dense and persona-sparse datasets, as will be seen in our experiments.

\begin{figure*}[ht]
\centering
\includegraphics[width=0.985\linewidth]{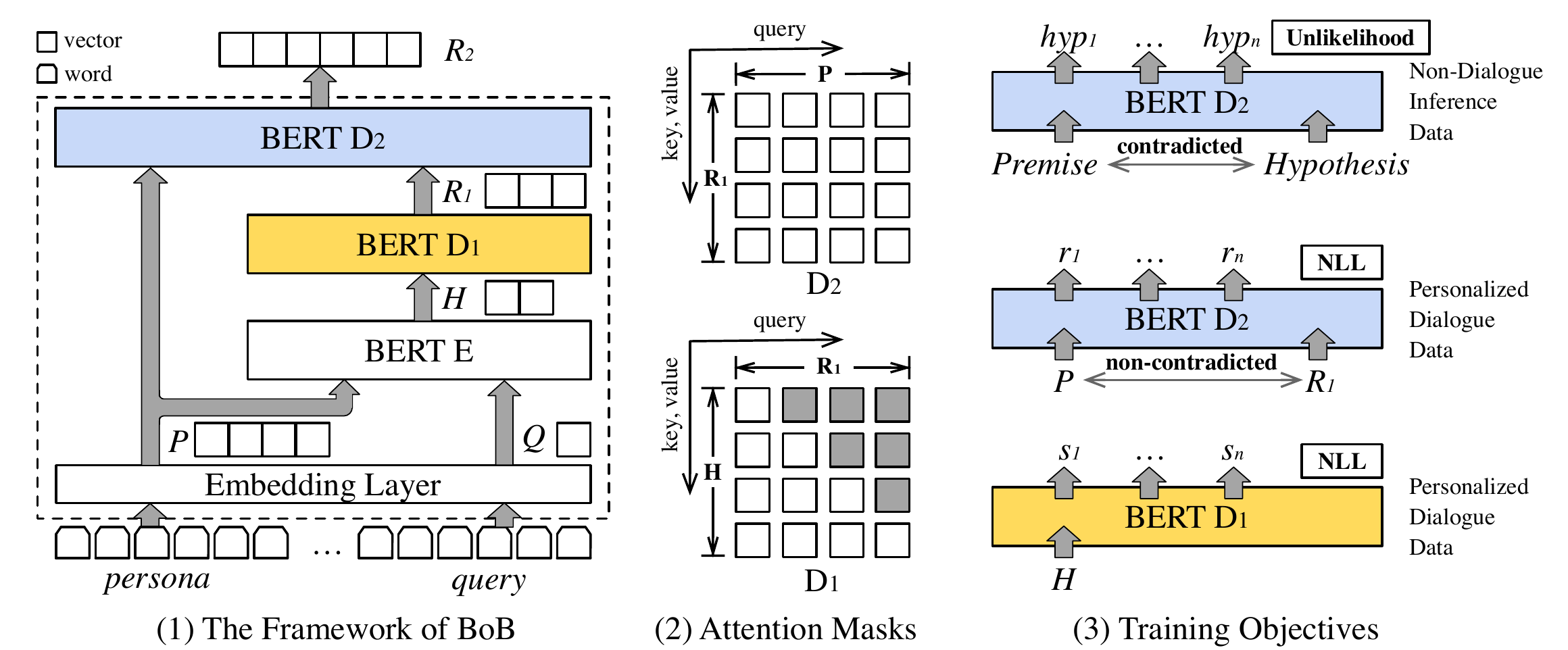}
\caption{ (1) The framework of the proposed BoB model, including an encoder (BERT $\mathbb{E}$), a response generation decoder (BERT $\mathbb{D}_1$), and a consistency understanding decoder (BERT $\mathbb{D}_2$). The italics denote the inputs and outputs of each submodule. (2) Transformer attention masks for generation ({$\mathbb{D}_1$}) and understanding ({$\mathbb{D}_2$}), and dark square means no attention. (3) Training objectives and the utilized data. NLL denotes negative log-likelihood.
}
\label{fig:overall_model}
\end{figure*}

\section{Model}

\subsection{Overview}
In this work, our goal is to learn a persona-based dialogue model from limited personalized data. To address the challenges of consistency understanding brought by limited data, we leverage large-scale non-dialogue inference data in our model.

Formally, let $\mathcal Q$ = $q_1,q_2,...,q_n$ denote the dialogue query, $\mathcal R$ = $r_1,r_2,...,r_m$ denote the target response, and $\mathcal P$ denote the personas. In addition, let $\mathcal N$ denote the non-dialogue inference data, which consists of premise, hypothesis, and their label. The premise and hypothesis are both natural sentences.
Note that in the following sections, we use fonts to distinguish between sentences ($\mathcal P$, $\mathcal Q$, $\mathcal R$) and their vector representations ($P$, $Q$, $R_1$, $R_2$).

The task of the proposed model $\mathbb M$ is to generate a persona consistent response $\mathcal{\hat R}=\hat r_1,\hat r_2,...,\hat r_m$, based on both persona $\mathcal P$ and query $\mathcal Q$, i.e., $\mathcal{\mathcal{\hat R}}={\mathbb M}(\mathcal Q,\mathcal P)$.
As shown in Figure~\ref{fig:overall_model}, the proposed model $\mathbb M$ consists of three BERT-based submodules: an encoder {$\mathbb E$}, a response decoder $\mathbb{D}_1$, and a consistency understanding decoder $\mathbb{D}_2$.
More concretely, {$\mathbb E$} encodes the embeddings of persona and query, i.e., $P$ and $Q$, into hidden states $H$. {$\mathbb{D}_1$} performs cross-attention on $H$ in a typical encoder-decoder manner, and generate a coarse representation $R_1$. {$\mathbb{D}_2$} learns consistency understanding from non-dialogue inference data $\mathcal N$ and further converts $P$ and $R_1$ into final representations $R_2$. At last, a consistent response $\mathcal{\hat R}$ could be generated from $R_2$.

\subsection{Disentangling}

For response generation, a typical persona-based dialogue model needs persona $\mathcal P$ and dialogue query $\mathcal Q$ to generate a response. For consistency understanding, a model needs persona $\mathcal P$, response $\mathcal R$, and the consistency labels between $\mathcal P$ and $\mathcal R$. However, if we entangle generation and understanding, it is not easy to obtain sufficient annotated data that satisfy the format of \{$\mathcal P$, $\mathcal Q$, $\mathcal R$, Label\}.

Instead, in our model, we design the decoder $\mathbb{D}_2$ to disentangle generation and understanding, where $\mathbb{D}_2$ maps $R_1$, rather than $Q$, to $R_2$.
The key to ``disentangling'' is we can get $R_1$ without the participation of $Q$, as $R_1$ is the representation of $\mathcal R$. As a result, the mapping from $R_1$ to $R_2$ could be independent of $Q$. In this way, it becomes possible to 1) learn persona-based dialogue generation from \{$\mathcal P$, $\mathcal Q$, $\mathcal R$\}, i.e., the personalized data, and 2) learn consistency understanding from \{$\mathcal P$, $\mathcal R$, Label\}. 
Moreover, considering the limited amount of such annotated data, we could approximate \{$\mathcal P$, $\mathcal R$, Label\} by the abundant non-dialogue inference data $\mathcal N$=\{Premise, Hypothesis, Label\}, where $\mathcal P$ and $\mathcal R$ corresponds to the Premise and Hypothesis.

Given data $\mathcal P$ and $\mathcal R$, suppose $\mathbb{D}_2$ understands persona consistency, it should maximize the likelihood of generating $\mathcal R$ if $\mathcal R$ is not contradicted to $\mathcal P$. Otherwise, it should minimize the likelihood of generating $\mathcal R$. 
Motivated by this observation, we choose to apply unlikelihood training on $\mathbb{D}_2$ to make it understand consistency. The detailed training objectives will be provided in Sec~\ref{sec:objectives}.

\subsection{BERT-over-BERT}
\label{sec:bob}
\subsubsection{Encoder}
The encoder $\mathbb E$ works like a standard BERT model, which bidirectionally encodes the input embeddings to a sequence of hidden vectors, from which the downstream tasks will be performed on.

In our model, the input consists of persona $\mathcal P$ and dialogue query $\mathcal Q$. For persona, whether $\mathcal P$ is personal facts (e.g., ``I have two dogs'') or profiles (e.g., ``location: Seattle''), we could always convert it into a sequence of words. A special token is placed between persona sequence and dialogue query, and the input is formated as:
\begin{gather}
\label{formula:1}
  input = p^{(0)}_1,p^{(0)}_2,...,p^{(t)}_{u_t},[s],q_1,q_2,...,q_n
\end{gather}
Then the embedding layer will convert $input$ into representations. Following usual practice, the input representations are the sum of the corresponding token, type, and position embeddings, where the type embedding is 0 and 1 for persona and query, respectively. $\mathcal P$ and $\mathcal Q$ can also get their independent representations. The resulted representations are $P$ and $Q$, which could be jointly denoted as $emb$ = $e^{p}_1,e^{p}_2,...,e^{q}_{l}$, where $l$ is the maximum length of the $input$.

Once we get the input representations, encoder $\mathbb E$ will perform multi-head attetnion~\cite{vaswani2017attention} on the $emb$ to transform the embeddings into a sequence of hidden vectors $H$. The multi-head attetnion could be denoted as MultiHead(query, key, value), where scaled dot-product attention is performed on query, key, and value. There are $N$ identical layers in $\mathbb E$, for each layer:
\begin{gather}
\label{formula:2}
  h^{i+1} = \text{FNN}(\text{MultiHead}(h^{i},h^{i},h^{i})),
\end{gather}
where $h^{0}$ = $emb$, and FNN is a fully connected feed-forward network containing two linear transformations with a ReLU activation in between. $h^{N}$ is the final output of encoder $\mathbb E$, i.e., $H$.

\subsubsection{Response Generation Decoder}
The response generation decoder $\mathbb{D}_1$ is initialized from BERT to inherit its robust language model but works in an auto-regressive decoder manner. First, a cross-attention is inserted between $\mathbb E$ and $\mathbb{D}_1$ to pass the context information. Second, a left-to-right mask is applied to $\mathbb{D}_1$ to preserve the auto-regressive generation property.

As the cross-attention does not exist in the BERT model, it is randomly initialized and updated during training.
In the cross-attention, the query comes from the previous layer of $\mathbb{D}_1$, and the key and value come from $H$:
\begin{gather}
\label{formula:3}
  r_1^{i+1} = \text{FNN}(\text{MultiHead}(r_1^{i}, H, H)).
\end{gather}
This attention is similar to the typical encoder-decoder attention mechanism in sequence to sequence models~\cite{bahdanau2014neural}, which attends to all positions in the context representations $H$ according to the variations of $r_1$. 
In training, $r_1^{0}$ is initialized from the embeddings of the target response. At each generation step, future tokens in the target response should not be considered. Therefore, as shown in Figure~\ref{fig:overall_model}, a left-to-right mask is applied to $\mathbb{D}_1$ to ensure that the predictions can only depend on the known outputs. 

$\mathbb{D}_1$ also has $N$ identical layers. And the output of the last layer $r_1^{N}$, i.e., $R_1$, is further fed to $\mathbb{D}_2$.

\subsubsection{Consistency Understanding Decoder}
Like $\mathbb{E}$ and $\mathbb{D}_1$, the consistency understanding decoder $\mathbb{D}_2$ is also initialized from BERT, from where $\mathbb{D}_2$ initializes a good semantic representation for understanding tasks.

In each layer of $\mathbb{D}_2$, the multi-head attention is performed twice:
\begin{gather}
\label{formula:45}
  p^{i+1} = \text{FNN}(\text{MultiHead}(r_2^{i}, P, P)),\\
  r_2^{i+1} = \text{FNN}(\text{MultiHead}(p^{i+1}, R_1, R_1)).
\end{gather}
The resulted $r_2^{i+1}$ in each layer thus fuses information from both $P$ and $R_1$. The output of the last layer of $\mathbb{D}_2$ is the final representations $R_2$. With an output layer, e.g. linear layers, upon the $R_2$, we can get the generated response ${\mathcal {\hat R}}$.

\subsection{Training Objectives}
\label{sec:objectives}
We employ negative log-likelihood (NLL) loss and unlikelihood loss for dialogue generation and consistency understanding. A brief illustration is shown in the last column of Figure~\ref{fig:overall_model} and detailed descriptions will be provided in this section.

\paragraph{Response Generation}
In our model, the widely adopted negative log-likelihood loss is applied in the training. For $\mathbb E$ and $\mathbb{D}_1$, 
they read the persona $\mathcal P$ and dialogue query $\mathcal Q$ to predict the target response $\mathcal R$, which yields the raw representations $R_1$:
\begin{equation}
\begin{aligned}
\label{formula:6}
  \mathcal{L}^{\mathbb{D}_1}_{NLL} &= -log(p_\theta(\mathcal R|\mathcal P,\mathcal Q))\\
  &= -\sum_{i=1}^{|\mathcal R|}log(p_\theta(r_i|\mathcal P,\mathcal Q,\mathcal R_{<i})).
\end{aligned}
\end{equation}
The generation part in $\mathbb{D}_2$ is also trained by NLL. $\mathbb{D}_2$ reads persona embeddings $P$ and raw representations $R_1$ to predict the target response $\mathcal R$:
\begin{equation}
\begin{aligned}
\label{formula:7}
  \mathcal{L}^{\mathbb{D}_2}_{NLL} &= -log(p_\gamma(\mathcal R|P, R_1))\\
  &= -\sum_{i=1}^{|\mathcal R|}log(p_\gamma(r_i|P,R_1,\mathcal R_{<i})).
\end{aligned}
\end{equation}

\paragraph{Unlikelihood Training}
Given large-scale non-dialogue inference dataset, we collect positive data $\mathcal D^+$ from the entailed category and collect negative data $\mathcal D^-$ from the contradicted category:
\begin{gather}
\label{formula:8}
  \mathcal D^+ = \{(\mathcal{\bar P}^{(i)},\mathcal{\bar R}^{(i)+})\},\ \ D^- = \{(\mathcal{\bar P}^{(j)},\mathcal{\bar R}^{(j)-})\},
\end{gather}
where $\mathcal{\bar P}$ and $\mathcal{\bar R}$ are premise and hypothesis from the non-dialogue inference data, and their representations in our model are denoted as $\bar P$ and $\bar R$. For data from $\mathcal D^+$, we still apply the NLL loss:
\begin{equation}
\begin{aligned}
\label{formula:9}
  \mathcal{L}^{\mathbb{D}^+_2}_{UL} = -\sum_{i=1}^{|\mathcal {\bar R}|}log(p_\gamma(\bar r_i|\bar P,\bar R,\mathcal {\bar R}_{<i})),
\end{aligned}
\end{equation}
For data from $\mathcal D^-$, we apply the unlikelihood objective to minimize the likelihood of contradictions:
\begin{equation}
\begin{aligned}
\label{formula:10}
  \mathcal{L}^{\mathbb{D}^-_2}_{UL} = -\sum_{i=1}^{|\mathcal {\bar R}|}log(1-p_\gamma(\bar r_i|\bar P,\bar R,\mathcal {\bar R}_{<i})),
\end{aligned}
\end{equation}
which penalizes every token in the contradicted target. Therefore, the loss $\mathcal{L}^{\mathbb{D}^-_2}_{UL}$ makes generating contradicted responses less likely.

\paragraph{Training Procedure}
The training steps can be summarized as follows:

1) Response Generation. Given $\mathcal P$, $\mathcal Q$, and $\mathcal R$ from personalized dialogue data, we calculate the response generation loss $\mathcal{L}_{1}=\mathcal{L}^{\mathbb{D}_1}_{NLL}+\alpha \mathcal{L}^{\mathbb{D}_2}_{NLL}$;

2) Consistency Understanding. Given $\mathcal D^+$ and $\mathcal D^-$ from non-dialogue inference data, we calculate the unlikelihood loss $\mathcal{L}_{2}=\beta\mathcal{L}^{\mathbb{D}_2^+}_{UL}+(1-\beta) \mathcal{L}^{\mathbb{D}_2^-}_{UL}$;

3) Optimization. Sum up $\mathcal{L}_{1}$ and $\mathcal{L}_{2}$. Update parameters with back-propagation.

We initialize our model from the publicly available BERT base model, with 12 layers and hidden size 768. We employ an Adam optimizer with a learning rate of varying from 5e-6 to 5e-5. Empirically, we set $\alpha$ to 5e-3 and $\beta$ to 0.1. The training of the proposed model was done on an Nvidia Telsa V100 32G GPU. Other details please refer to the released projects.

\section{Experiments}

\subsection{Datasets}

To evaluate the performance of the proposed model, we carried out persona-based dialogue generation experiments in a persona-dense scenario and a persona-sparse scenario with two publicly available datasets:
\begin{itemize}
  \item {\bf PersonaChat}~\cite{zhang-etal-2018-personalizing} is a crowd-sourced dataset covering rich persona features. The dialogues in this dataset are grounded on specific personal facts. Here we use the ConvAI2 PersonaChat~\cite{dinan2019second}, so the results are comparable to existing methods.
  \item {\bf PersonalDialog}~\cite{Zheng2019PersonalizedDG} is a large-scale persona-sparse dataset, which is collected from Chinese social media Weibo. This dataset provides persona profiles and dialogues, but the majority of the dialogues are not persona-related. Two testsets are provided: a random testset, which is identically distributed as the training data, and a biased testset, which is manually selected to cover persona-related features.
\end{itemize}
We summarize the key statistics of two personalized dialogue datasets in Tabel~\ref{tab:datasets}. 

As aforementioned, we leverage non-dialogue inference data to address the consistency understanding issue brought by limited personalized data. Here we use the non-dialogue inference dataset MNLI~\cite{williams-etal-2018-mnli} and its Chinese version CMNLI~\cite{xu-etal-2020-clue} as our auxiliary data. Moreover, to better compare models' performance on persona consistency, we leverage two dialogue inference datasets, DNLI~\cite{welleck-etal-2019-dialogue} and KvPI~\cite{song-etal-2020-kvpi}, for evaluations.
The statistics\footnote{Note that for the DNLI, we only count the tuples that can be restored as \{persona, query, response\} in our experiments.} of these inference datasets are summarized in Table\ref{tab:inference_datasets}.

\subsection{Compared Methods}

The following models, including both non-pretrained and pretrained ones, have been compared in the experiments.

\paragraph{Baselines.} Vanilla {\bf Transformer}~\cite{vaswani2017attention} is employed as baselines for the experiments on both PersonaChat and PersonalDialog. 
Personas are concatenated to the dialogue queries.

\paragraph{Non-Pretrained Models.} Meta-learning has recently been explored in addressing the limited personalized data issue.
{\bf CMAML}~\cite{song-etal-2020-learning} is a meta-learning based method that learns from few shot personas by customizing the model structures. Besides the meta-learning methods, {\bf GDR}~\cite{song-etal-2020-generate} introduces inference ability on the PersonaChat with a generate-refine framework. However, the two models are elaborately designed for the persona-dense dataset and not appliable for the persona-sparse scenario. Thus we only employ them for experiments on PersonaChat.

\paragraph{Pre-training Models.}
In the ConvAI2 challenge~\cite{dinan2019second}, which utilizes PersonaChat as the competition dataset, {\bf LIC}~\cite{golovanov2020lost} is the best performing model. Thus we compare this model in the experiments on both PersonaChat and PersonalDialog. {\bf AttentionRouting}~\cite{Zheng_Zhang_Huang_Mao_2020} is a pre-training method specially designed for the persona-sparse dataset, and it is also the latest model on PersonalDialog. We also finetune a {\bf GPT2}~\cite{radford2019language} for a thorough comparison on PersonaChat.

\begin{table}[t]
\centering
\resizebox{\columnwidth}{!}{%
\begin{tabular}{llll}
\toprule
\multicolumn{1}{c}{\textbf{Dataset}} &
  \multicolumn{1}{c}{\textbf{\# Train}} &
  \multicolumn{1}{c}{\textbf{\# Valid}} &
  \multicolumn{1}{c}{\textbf{\# Test}} \\ \midrule
PersonaChat     & 121,880           & 9,558             & 7,801            \\
PeronalDialog    & 5,014,349         & 423,817           & 10,000 / 521     \\ \bottomrule
\end{tabular}%
}
\caption{Statistics of persona-based dialogue datasets.}
\label{tab:datasets}
\end{table}

\begin{table}[t]
\centering
\resizebox{.91\columnwidth}{!}{%
\begin{tabular}{llll}
\toprule
\multicolumn{1}{c}{\textbf{Dataset}} &
  \multicolumn{1}{c}{\textbf{\# Entailed}} &
  \multicolumn{1}{c}{\textbf{\# Neutral}} &
  \multicolumn{1}{c}{\textbf{\# Contra.}} \\ \midrule
MNLI  & 130,615 & 130,590 & 130,590 \\
CMNLI & 130,612 & 130,555 & 130,616 \\ \hline
DNLI  & 15,495  & 20,927  & 16,488  \\
KvPI  & 33,114  & 54,426  & 31,000  \\ \bottomrule
\end{tabular}%
}
\caption{Statistics of different inference datasets.}
\label{tab:inference_datasets}
\end{table}

\subsection{Evaluation Metrics}
We focus on two main aspects of the persona-based dialogues: {\bf response quality} and {\bf persona consistency}. To compare different models, we employ both automatic metrics and human evaluations.

\paragraph{Automatic Metrics} For dialogue quality, we employ perplexity ({\bf PPL.}) and distinct 1/2 ({\bf Dist.1/2}) following common practice~\cite{zhang-etal-2018-personalizing,Zheng_Zhang_Huang_Mao_2020}. Lower perplexity means better language modeling. Distinct 1/2~\cite{li-etal-2016-diversity} are the ratio of distinct uni-grams / bi-grams, and higher distinct means better reponse diversity.

For persona consistency, we employ two metrics. The first is Consistency Score ({\bf C.Score})~\cite{madotto-etal-2019-personalizing}, which leverages a referee model to predict consistency and can be defined as:
\begin{equation}
\label{formula:11}
\begin{aligned}
\text{NLI}(r,p_i)&=\left\{
\begin{aligned}
-1 & , & \text{if}\ r\ \text{contradicts}\ p_i, \\
0 & , & \text{if}\ r\ \text{is irrelevant to}\ p_i, \\
1 & , & \text{if}\ r\ \text{entails}\ p_i.
\end{aligned}
\right.\\
\text{C.Score}(r) &= \sum\nolimits_{i=1}^t\text{NLI}(r,p_i).
\end{aligned}
\end{equation}
Here the NLI is a pre-trained RoBERTa model~\cite{liu2019roberta} finetuned with the dialogue inference datasets, i.e., DNLI and KvPI, as descriped in Table~\ref{tab:inference_datasets}. The RoBERT model achieves testset accuracy of 89.3\% and 88.9\% on DNLI and KvPI, which is aligned to the reported 88.20\%~\cite{welleck-etal-2019-dialogue} and 88.0\%~\cite{song-etal-2020-kvpi}.

The second metric is Delta Perplexity ({$\Delta$}{\bf P}), which evaluates consistency from model's internal distributions.
\citet{li-etal-2020-dont} first calculates the perplexity of entailed ({\bf p.Ent}) and contradicted ({\bf p.Ctd}) dialogues in the inference dataset. A dialogue model with good understanding ability should assign lower perplexity to the entailed dialogues while higher perplexity to the contradictions. From this intuition, the {$\Delta$}{P} can be defined as:
\begin{equation}
\label{formula:12}
\begin{aligned}
{\Delta \text P} = \text{PPL}(\text{Contradicted}) - \text{PPL}(\text{Entailed}),
\end{aligned}
\end{equation}
where a larger {$\Delta$}{P} means the model has a better ability to distinguish entailment from contradiction.
In our experiments, we get entailed and contradicted \{persona, query, response\} tuples from the dialogue inference datasets DNLI and KvPI.

\begin{table*}[ht]
\resizebox{\textwidth}{!}{%
\begin{tabular}{@{}r|llllllll|llll@{}}
\toprule
 &
  \textbf{PPL} &
  \textbf{Dist.1} &
  \textbf{Dist.2} &
  \textbf{D.AVG} &
  \textbf{p.Ent} &
  \textbf{p.Ctd} &
  \textbf{$\Delta$P} &
  \textbf{C.Score} &
  \textbf{Flue.} &
  \textbf{Info.} &
  \textbf{Relv.} &
  \textbf{Per.C.} \\ \midrule
Transformer & 28.8 & 3.14 & 17.80 & 10.47 & 31.5 & 35.5 & 4.0  & 1.20  & 3.05 & 2.57 & 2.72 & 0.05 \\
CMAML       & 36.7 & 1.00 & 2.10  & 1.55  & 32.3 & 37.5 & 5.2  & 6.96  & 3.36 & 2.40 & 3.09 & 0.24 \\
GDR         & 16.7 & 3.76 & 23.10 & 13.43 & 19.7 & 32.3 & 12.6 & 7.89  & 3.38 & 2.74 & 3.13 & 0.21 \\
LIC         & 17.3 & 6.29 & 28.99 & 17.64 & 13.7 & 20.4 & 6.7  & 14.12 & 3.70 & 3.53 & 3.47 & 0.39 \\
GPT2        & 14.4 & 7.29 & 28.12 & 17.71 & 12.0 & 20.2 & 8.2  & 15.88 & 3.79 & 3.22 & 3.79 & 0.47 \\ \midrule
BoB (Ours) &
  \textbf{7.8} &
  \textbf{8.40} &
  \textbf{36.08} &
  \textbf{22.24} &
  \textbf{7.3} &
  \textbf{83.4} &
  \textbf{76.1} &
  \textbf{17.18} &
  \textbf{4.12} &
  \textbf{4.03} &
  \textbf{4.09} &
  \textbf{0.60} \\ \bottomrule
\end{tabular}%
}
\caption{Automatic and human evaluation results on the full PersonaChat dataset. The best results are in bold.}
\label{tab:dense-full}
\end{table*}

\begin{table*}[ht]
\resizebox{\textwidth}{!}{%
\begin{tabular}{@{}r|llllllll|llll@{}}
\toprule
 &
  \textbf{PPL} &
  \textbf{Dist.1} &
  \textbf{Dist.2} &
  \textbf{D.AVG} &
  \textbf{p.Ent} &
  \textbf{p.Ctd} &
  \textbf{$\Delta$P} &
  \textbf{C.Score} &
  \textbf{Flue.} &
  \textbf{Info.} &
  \textbf{Relv.} &
  \textbf{Per.C.} \\ \midrule
Baselines' Best & 14.4 & 7.29 & 28.99 & 17.71 & 12.0 & 37.5 & 12.6 & 15.88 & 3.79 & 3.53 & 3.79 & 0.47 \\ \midrule
Ours 1/8 Data   & 11.6$^\dagger$ & 7.49$^\dagger$ & 27.10 & 17.30 & 11.3$^\dagger$ & 83.6$^\dagger$ & 72.3$^\dagger$ & 15.87 & 4.17$^\dagger$ & 3.48 & 4.12$^\dagger$ & 0.62$^\dagger$ \\
Ours 1/4 Data   & 9.7  & 7.97 & 30.20$^\dagger$ & 19.09$^\dagger$ & 11.8 & 85.8 & 74.0 & 16.04$^\dagger$ & 4.19 & 3.47 & 4.17 & 0.60 \\
Ours 1/2 Data   & 8.9  & 8.13 & 33.08 & 20.61 & 8.1  & 81.9 & 73.8 & 16.36 & 4.03 & 3.70$^\dagger$ & 3.94 & 0.61 \\ \bottomrule
\end{tabular}%
}
\caption{Automatic and human evaluation results of the low resource settings on the PersonaChat dataset. The $\dagger$ means the minimum amount of data our model needed to outperform baselines' best results. }
\label{tab:dense-low-resource}
\end{table*}

\paragraph{Human Evaluations}
We recruit two teams (one for English and another for Chinese), each consists of five professional annotators, from a third-party company. These annotators are proficient in language tasks but know nothing about the models. We sample 100 \{persona, query, response\} tuples for each model's evaluation under every setting.

Human annotators are asked to evaluate dialogue quality from three conventional criteria: fluency ({\bf Flue.}), informativeness ({\bf Info.}), and relevance ({\bf Relv.}). Each criterion is rated on a five-scale, where 1, 3, and 5 indicate unacceptable, moderate, and perfect performance, respectively. The annotators are also instructed to label the consistency ({\bf Per.C.}) between persona and response, where 1 means persona-related and consistent, 0 means irrelevant, and -1 means contradicted.

\subsection{Persona-Dense Results}

\paragraph{Full PersonaChat}
We first report the full PersonaChat experimental results in Table~\ref{tab:dense-full}. Our method achieves better performance consistently across all automatic and human evaluation metrics, which shows the effectiveness of our model.

Among all the metrics, our model obtains significant improvements on PPL and $\Delta$P. The lowest testset PPL means our model has learned a good language model fitting this dataset. Moreover, the highest $\Delta$P shows that our model could more effectively distinguish entailment from contradiction than other baselines, which indicates our model has a better understanding of persona consistency.

\paragraph{Less Personalized Data}
Now that our model achieves better performance with a large margin on the full PersonaChat dataset, we want to test our model by simulating a low-resource scenario~\cite{zhao2019low}, where we gradually reduce the number of examples by halving the training set. We report the low-resource settings' results in Table~\ref{tab:dense-low-resource}.

As we can see, our model can outperform most of the baselines' best results even by using only 1/8 of the training data. The performance gains largely benefit from the powerful language model of the backbone BERT model. Furthermore, due to the disentangling of generation and understanding, our model presents a stable performance on $\Delta$P regardless of the size of the training set. This is in line with our expectations because the proposed model learns consistency understanding from the non-dialogue inference data rather than the persona-dense dialogue data.
We observe that the method also improves fluency and informativeness. It is mainly due to the introduction of the non-dialogue inference data in the training procedure, which potentially enriches the dialogue language model.

\subsection{Validations on Persona-Sparse}
We further validate our model on a persona-sparse scenario. To have a more intuitive understanding of ``sparsity'', we recruit the same annotation team to annotate whether the dataset response is persona-related in the sampled random and biased test data. Results show that only 1\% responses are persona-related in the random test data and 28\% in the biased test data. We calculate the Fleiss’ Kappa among the five annotators and obtain a kappa of 0.774, which means {\it substantial agreement}~\cite{landis1977measurement}.
We report the evaluation results on both random and biased testsets in Table~\ref{tab:sparse-main}.

On the random test set, experimental results demonstrate that our model has some advantages over other methods, but no method can consistently outperform the others. One possible reason is that the task has degenerated into the ordinary dialogue generation in the random test set, so our model's advantages can not be effectively leveraged. In contrast, on the biased test set, our model achieves the best performance on most metrics. 
The good performance on the metrics C.Score and Per.C. indicates that our model can be effectively trained from a dataset with limited personalized dialogues.

\begin{table*}[ht]
\resizebox{\textwidth}{!}{%
\begin{tabular}{@{}r|llllll|llllll|l@{}}
\toprule
       & \multicolumn{6}{c|}{\textbf{Random Testset}}                       & \multicolumn{6}{c|}{\textbf{Biased Testset}}      & \multicolumn{1}{c}{\textbf{KvPI}} \\ \cmidrule(l){2-14} 
 &
  \textbf{PPL} &
  \textbf{C.Score} &
  \textbf{Flue.} &
  \textbf{Info.} &
  \textbf{Relv.} &
  \textbf{Per.C.} &
  \textbf{PPL} &
  \textbf{C.Score} &
  \textbf{Flue.} &
  \textbf{Info.} &
  \textbf{Relv.} &
  \textbf{Per.C.} &
  \textbf{$\Delta$P} \\ \midrule
Trans  & 43.7 & 0.95          & 3.26 & 2.38 & 2.72          & 0.00          & 83.2 & 1.04  & 3.54 & 2.58          & 2.84 & 0.03 & 3.28                              \\
LIC    & 47.8 & \textbf{4.08} & 3.68 & 2.66 & 2.92          & \textbf{0.02} & 43.3 & 8.25  & 3.72 & 3.01          & 3.04 & 0.08 & 2.86                              \\
AR     & 34.2 & -2.14         & 3.71 & 2.58 & \textbf{3.02} & -0.03         & 38.7 & 11.72 & 3.78 & 3.11          & 3.10 & 0.13 & 3.08                              \\ \midrule
Ours &
  \textbf{18.5} &
  2.10 &
  \textbf{3.75} &
  \textbf{2.69} &
  2.98 &
  0.01 &
  \textbf{19.5} &
  \textbf{12.76} &
  \textbf{3.84} &
  3.13 &
  \textbf{3.17} &
  \textbf{0.15} &
  \textbf{85.40} \\ \midrule
w/o UL & 19.3 & -3.13         & 3.73 & 2.57 & 2.93          & -0.06         & 20.1 & 10.53 & 3.79 & 2.92          & 3.10 & 0.09 & 4.10                              \\
E+D1   & 31.7 & 0.15          & 3.74 & 2.68 & 2.96          & -0.01         & 38.0 & 9.75  & 3.74 & \textbf{3.15}    & 3.06 & 0.08 & 2.80                              \\
E      & 35.5 & 1.64          & 3.67 & 2.57 & 2.96          & 0.01          & 41.1 & 7.41  & 3.72 & 3.05 & 3.04 & 0.04 & 4.60                              \\ \bottomrule
\end{tabular}%
}
\caption{Automatic and human evaluation results on the random testset and biased testset of PersonalDialog, along with the ablation results. Trans denotes Transformer, and AR denotes AttentionRouting. Best results in bold.}
\label{tab:sparse-main}
\end{table*}

\begin{table}[ht]
\resizebox{\columnwidth}{!}{%
\begin{tabular}{@{}r|ll|llll@{}}
\toprule
 & \textbf{PPL} & \textbf{$\Delta$P} & \textbf{Flue.} & \textbf{Info.} & \textbf{Relv.} & \textbf{Per.C.} \\ \midrule
Ours    & 7.8  & 76.1 & 4.12 & 4.03 & 4.09 & 0.60 \\ \midrule
w/o UL & 8.1  & 7.8  & 3.81 & 3.50 & 3.80 & 0.48 \\
E+D$_1$   & 23.6 & 4.9  & 3.65 & 3.18 & 3.60 & 0.45 \\
E      & 25.7 & 7.1  & 3.69 & 3.28 & 3.60 & 0.42 \\ \bottomrule
\end{tabular}%
}
\caption{Ablation results of automatic metrics and human evaluations with full PersonaChat dataset.}
\label{tab:dense-ablation}
\end{table}

\subsection{Analysis and Ablation Study}

In addition to the good performance of the BoB model, we are also curious about 
{\bf Q1}: what is the key to the BoB model's understanding ability?
{\bf Q2}: can the pre-trained models understand persona consistency just through finetuning on the personalized dialogues?
And {\bf Q3}: does the extremely low PPL come from the initialization of the BERT model or the architecture of the proposed BoB model?

To better answer the above questions, we ablate the BoB model in the following three ways: 
1) {\bf w/o UL}. It removes the unlikelihood objective. 
2) {\bf ${\mathbf E}$+${\mathbf{D}_1}$}. It removes the unlikelihood objective and the second decoder $\mathbb D_2$.
3) {\bf ${\mathbf E}$}. It removes the unlikelihood objective and both decoders and thus degenerates into a vanilla BERT model.
We report the ablation results on PersonalDialog in Table~\ref{tab:sparse-main} and full PersonaChat in Table~\ref{tab:dense-ablation}. From these results:

{\bf Answer to Q1:}
The key to our model's understanding is the unlikelihood training. In training, our model assigns large perplexity to the contradictions. In generating, the non-contradicted responses are more likely to be generated as they are with much smaller losses. Table~\ref{tab:cases} shows an example.
And as presented in the results, after removing the unlikelihood objective,
all ablated models suffer from significant performance degradations in consistency-related metrics, such as Per.C. and $\Delta$P.

{\bf Answer to Q2:}
Pretrained models barely understand consistency from personalized dialogues. According to the poor performances on $\Delta$P, the three BERT-based ablated models can hardly distinguish contradiction from entailment. Although their Per.C. metric still looks good, it may come from just mimicking and copying words rather than understanding. A similar phenomenon also occurs to the pre-trained GPT2, as shown in Table~\ref{tab:dense-full}. 
It is also this phenomenon that motivates us to introduce the unlikelihood training into the BoB model.

{\bf Answer to Q3:}
$\mathbb D_2$ in the BoB architecture contributes most to the PPL. As shown in both datasets' ablation results, the PPL decreases the most after removing $\mathbb D_2$. We can also see an apparent gap between the models with $\mathbb D_2$ and the vanilla BERT on PPL. Nevertheless, the BERT model still offers a good initialization for the BoB model to achieve the best performance on different metrics.

\begin{table}[]
\resizebox{\columnwidth}{!}{%
\begin{tabular}{@{}r|l@{}}
\toprule
\textbf{Persona} & I've a son who is in junior high school                      \\
\textbf{Query}   & You have any children?                                       \\ \midrule
\textbf{GPT2}   & No kids. I work at home depot so I'm busy.                   \\
\textbf{Ours}    & Yes, I have a son in the 8th grade.                         \\ \bottomrule
\end{tabular}%
}
\caption{A generated example from our model.}
\label{tab:cases}
\end{table}

\subsection{Reproducibility}
The implementation for the BoB model is released at https://github.com/songhaoyu/BoB.

\section{Conclusions}
In this work, we propose a novel BERT-based dialogue model to learn from limited personalized data by disentangling response generation and consistency understanding. Unlikelihood training with non-dialogue inference data is introduced to enhance the model's understanding ability.
Experiments on two publicly available datasets demonstrate that our model can be trained with limited personalized dialogue data while still obtain significant improvements over strong methods.

\section*{Acknowledgments}

This paper is supported by the National Natural Science Foundation of China under Grant No.62076081, No.61772153, and No.61936010, and supported by the Science and Technology Innovation 2030 Major Project of China under Grant No.2020AAA0108605. We thank all the anonymous reviewers for their helpful comments and suggestions.

\section*{Ethical Statement}
Persona-based dialogue research intends to address the persona inconsistency issue in open-domain dialogue to facilitate human-computer interactions. Giving dialogue system a specific persona is a mainstream to alleviate the inconsistency issue of dialogues under the current stage. The purpose is to endow the dialogue system with self logical consistency rather than imitate specific human beings. 
Simultaneously, in this work, the data resources we use are all from published works and do not involve privacy issues related to data collection. We also confirm that this work neither automatically infers or attributes identity characteristics to the participants nor categorizes them in the training datasets.

\bibliographystyle{acl_natbib}
\bibliography{acl2021}


\end{document}